\documentclass[11pt,a4paper]{article}
\usepackage[hyperref]{emnlp2018}
\usepackage{times}
\usepackage{latexsym}
\usepackage[english]{babel}
\usepackage{url}
\usepackage{natbib}
\usepackage{graphicx}
\usepackage{bm}
\usepackage{multirow}
\usepackage{pifont}
\usepackage{amsmath} 
\usepackage{amssymb}
\usepackage{amsthm}
\usepackage{algorithm}
\usepackage{algpseudocode}
\usepackage{varwidth}

\newtheorem{theorem}{Theorem}

\algnewcommand{\LeftComment}[1]{\Statex \(\triangleright\) #1}
\theoremstyle{definition}

\usepackage[inline]{enumitem}
\usepackage[toc,page]{appendix}
\usepackage{caption}
\usepackage[export]{adjustbox}
\usepackage[para]{footmisc}

\aclfinalcopy 

\title{Effective Context and Fragment Feature Usage \\ for Named Entity Recognition}
  
  \author{Nargiza Nosirova, Mingbin Xu, Hui Jiang \\
  Department of Electrical Engineering and Computer Science \\
  Lassonde School of Engineering, York University \\
  4700 Keele Street, Toronto, Ontario, Canada \\
  {\tt \{nana, xmb, hj\}@cse.yorku.ca}
 }

\date{}

\begin{document}
\maketitle
\begin{abstract}
In this paper, we explore a new approach to named entity recognition (NER) with the goal of learning from context and fragment features more effectively, contributing to the improvement of overall recognition performance. We use the recent fixed-size ordinally forgetting encoding (FOFE) method to fully encode each sentence fragment and its left/right contexts into a fixed-size representation. Next, we organize the context and fragment features into groups, and feed each feature group to dedicated fully-connected layers. Finally, we merge each group's final dedicated layers and add a shared layer leading to a single output. The outcome of our experiments show that, given only tokenized text and trained word embeddings, our system outperforms our baseline models, and is competitive to the state-of-the-arts of various well-known NER tasks.
\end{abstract}

\section{Introduction}
Named entity recognition is the task of identifying proper nouns in a given text, and categorizing them into various types of entities. 
It is a fundamental problem in NLP, and its usefulness extends to tasks such as summarization and question answering \cite{aramaki, ravichandran}.  Traditional NER methods involve using hand-crafted features, such as conditional random fields (CRFs) . For example, \newcite{mccallum} use a CRF model with a web-based lexicon as a feature enhancement, while \newcite{che} and \newcite{krishnan} show the benefits of using non-local features. Over the recent years, researchers have turned to neural network architectures using non hand-crafted features. 
For example, \newcite{collobert2011} proposed a neural architecture that learns from word embeddings and requires little feature engineering.
However, in his use of feed-forward neural networks (FFNNs), the context used around a word is restricted to a fixed-size window, which could result in the loss of potentially relevant information between words that are further apart. 
 \newcite{xu} has recently proposed a non-sequence labelling method for NER with FOFE features, which can encode any variable-length sequence of words into a fixed-size representation. This method alleviates the limitations of Collobert's \shortcite{collobert2011} FFNN model, since the encoding uses the whole context around a word within the sentence, without settling for a fixed-size window.
Our main contribution lies in extending the model suggested by \newcite{xu}. In this paper, we propose a FOFE-based neural network model dedicating separate initial layers for fragment and context features and merging them into a shared layer to perform a unified prediction. Experimental results have shown that this method yields competitive results compared to the state-of-the-arts while increasing recall compared to our baseline models. 
\section{Model}
Our neural network model is inspired by the work of \newcite{xu}, where we use a local detection approach relying on the FOFE method to fully encode each sentence fragment and its contexts. Instead of using consecutive fully-connected layers that handle both context and fragment features, we propose to dedicate the initial fully-connected layers of the network to each feature kind, and subsequently combine the layers into a single shared layer that leads to a single output. 
\subsection{Fixed-Size Ordinally Forgetting Encoding (FOFE) }
In this section, we describe the FOFE method. Given a vocabulary $V$, each word can be represented by a 1-of-$|V|$ one-hot vector.  FOFE mimics bag-of-words but incorporates a forgetting factor to capture positional information. It encodes any variable length sequence composed of words in $V$. Let $S = w_{1} \cdots  w_{N}$ denote a sequence of $N$ words from $V$, and denote $\bm{e_n}$ to be the one-hot vector of the $n$-{th} word in $S$, where $1 \leq n \leq N$. Assuming  $\bm{z_0} = \mathbf{0}$,  FOFE generates the code using a simple recursive formula from word $w_{1}$ to $w_{n}$ of the sequence as follows:
\begin{equation*}
\bm{z_n} = \alpha \cdot \bm{z_{n - 1}} + \bm{e_n}  \label{eq_FOFE_formula}
\end{equation*}
where $\alpha$ is a constant forgetting factor. Hence, $\bm{z_n}$ can be viewed as a fixed-size representation of the subsequence $\{ w_{1}, w_{2}, \cdots, w_{n} \}$. 

The theoretical properties that show FOFE code uniqueness are as follows:
\begin{theorem}
If the forgetting factor $\alpha$ satisfies  $0 < \alpha \le 0.5$, FOFE is unique for any countable vocabulary $V$ and any finite value $N$ .
\end{theorem}

\begin{theorem}
For $0.5 < \alpha < 1$, given any finite value $N$ and any countable vocabulary $V$, FOFE is almost unique everywhere, except only a finite set of countable choices of $\alpha$. 
\end{theorem}

When $0.5 < \alpha < 1$, uniqueness is not guaranteed. However, the odds of ending up with such scenarios is small. Furthermore, it is rare to have a word reappear many times within a near context. Thus, we can say that FOFE can uniquely encode any sequence of variable length, providing a fixed-size lossless representation for any sequence. The proof for those theorems can be found in \newcite{zhanghui}.

\subsection{FOFE Context \& Fragment Features} \label{feature_section}
\noindent
\subparagraph{Fragment Features}
At word level, we extract the bag-of-words of the sentence fragment in both cased and uncased forms. Since we can view the fragment as a cased character sequence, it can be encoded with FOFE. We encode the sequence from left to right as well as from right to left. The encodings are then projected into a trainable character embedding matrix. For a fair comparison, we also use character CNNs to generate additional character-level features \cite{kim}.
\noindent
\subparagraph{Context Features}
We convert the contexts of the fragment within the sentence to FOFE codes at word-level in cased and uncased forms, once containing the fragment, and once without. Those codes are then projected to lower-dimensional dense vectors using projection matrices. Those projection matrices are pre-trained using word2vec \cite{mikolov} and allowed to be modified during training.
\begin{figure}[t]
	\centering
\includegraphics[width=0.5\textwidth, center]{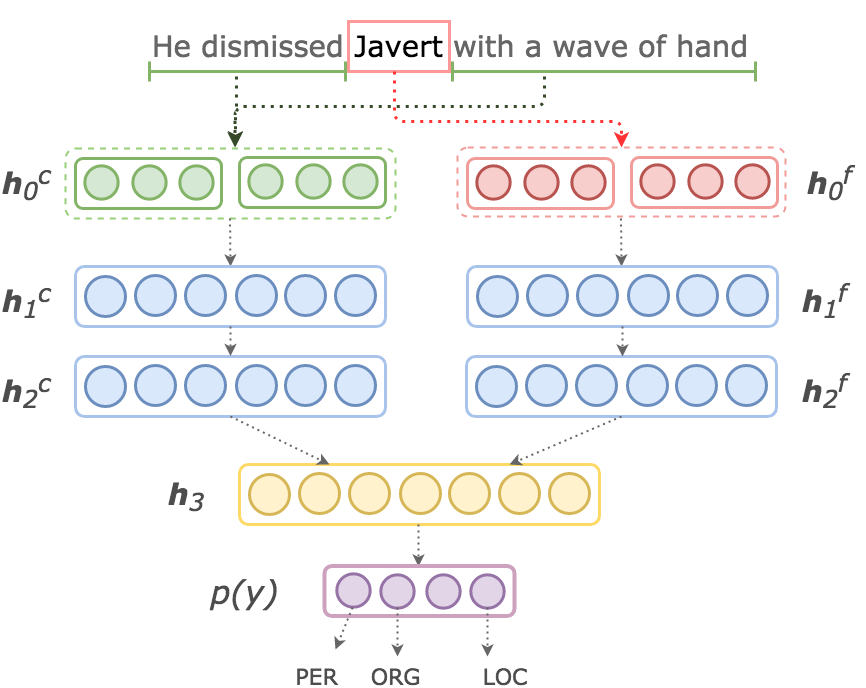}
\caption{Illustration of an example structure of out model using FOFE codes. The window currently examines the fragment {\it Javert}.}
\label{Fig:FOFE-NER-diagram}
\end{figure}
\subsection{Effective Context \& Fragment Feature Usage for NER}
We aim to consider influences between contexts and their corresponding fragments. If the context of a named entity fragment is not indicative of the fragment as being such, we can resort to learning the morphology of the fragment itself and grasp patterns that could lead us to believe that the fragment is indeed a named entity. By dedicating layers to each feature kind, we ensure that the context-based layer signals are tuned to identify entities based on the surrounding context, while the fragment-based layer signals identify them based on the morphology of the fragment itself. Since the layers merge into a shared layer, this permits the model to have a higher chance of predicting entities that would be hard to recognize based on the context, but self-evident based on the fragment. Furthermore, our model structure provides the flexibility of modeling information based on multiple sources of knowledge. 
Figure \ref{Fig:FOFE-NER-diagram} illustrates an example of our neural architecture:
\begin{enumerate}\itemsep0em
	\item The context and fragment features are extracted from the text based on section \ref{feature_section} and concatenated within their categories resulting in $\mathbf{h}_{0}^{w} \big| w \in \{ f, c \}$.
	\item  Two hidden layers $\mathbf{h}_{1}^{w}$ and $\mathbf{h}_{2}^{w}$ are fully connected to each category's embedding layer $\mathbf{h}_{0}^{w}$.
	\item A shared hidden layer $\mathbf{h}_{3}$ is fully connected to each $\mathbf{h}_{2}^{w}$. 
	\item The final layer is a softmax layer which outputs the probability distribution over classes, $\mathit{p}(y)$.
\end{enumerate}
Each layer $h_{j, j > 0}$ consists of ReLUs \cite{nairhinton} and are initialized based on a uniform distribution following \newcite{glorot}.

\begin{table*}[!htbp]
	\centering
	\scalebox{0.95}{
	\begin{tabular}{|l|lll|lll|lll|}
		\hline
		 \multirow{2}{*}{LANG}  &
			\multicolumn{3}{c|}{\newcite{xu}} & \multicolumn{3}{c|}{Our model} & \multicolumn{3}{c|}{2016 Best} \\
		 \ & P & R & F1 & P & R & F1 & P & R & F1 \\
		\hline \hline
		  	ENG & 0.836 & {\bf 0.680} & {\bf 0.750} & 0.812 & {\bf 0.756} &{\bf 0.782} & 0.846 & 0.710 & {\bf 0.772} \\
		   CMN & 0.789 & 0.625 & 0.698 & 0.797 & 0.693 & {\bf 0.741} & 0.789 & 0.737 & 0.762 \\
		   SPA & 0.835 & 0.602 & 0.700 & 0.848 & 0.608 & 0.708 & 0.839 & 0.656 & 0.736 \\
		   ALL & 0.819 & 0.639 & 0.718 & 0.815 & 0.693 & {\bf 0.749} & 0.802 & 0.704 & {\bf 0.756} \\
		\hline
	\end{tabular} }
	\caption{Comparison of our model to the baseline models in \newcite{xu} as well as to the best system for the KBP 2016 task.  }
	\label{tbl:kbp2016}
\end{table*}

\subparagraph*{Training} 
At each training step, we randomly choose a training sample represented as a one of the feature forms and forward pass. Next, we backpropagate the loss of the current instance through the shared and feature dedicated layers and update the model parameters. 
For predicting models relative to the ground truth, we use categorical cross entropy loss. For optimization, we use mini-batch SGD with momentum of 0.9 \cite{bottou} and learning rates decaying exponentially by a factor of $1/16$. 
The mini-batch size is set to $128$ for all experiments. Grid search is used for the other hyper-parameters, tuned against the task's development set with early stopping. The FOFE forgetting factor for all models are set to $\alpha_{w} = 0.5$ for words, and $\alpha_{c} = 0.8$ for characters. We apply dropout \cite{srivastava} to all layers with $0.5$ probability.
The same post-processing and decoding steps are followed as in \newcite{xu}.
Detailed hyper-parameter settings used in our experiments are given in Appendix A. 
\section{Experiments}
We experiment with four diverse NER tasks of different languages: CoNLL-2003 English, OntoNotes 5.0 English and Chinese, trilingual KBP 2016 (English, Chinese and Spanish), and CoNLL-2002 Spanish. For the CoNLL-2003 task, we use cased and uncased word embeddings of size $256$ trained on the Reuters RCV1 corpus. The remaining tasks use cased and uncased word embeddings of size $256$ trained on the English \cite{parker2011english}, Spanish \cite{mendonca2009spanish}  and Chinese \cite{graff2005chinese} Gigaword for the corresponding models evaluated in that language. 
\subparagraph{Dataset Description}

\textit{CoNLL-2003 ENG}: The CoNLL-2003 \cite{tjong2003} dataset consists of newswire data from the Reuters RCV1 corpus. It has four entity types: person, location, organization and miscellaneous. \textit{OntoNotes 5.0 ENG and ZH}: The OntoNotes dataset is built from sources such as broadcast conversation and news, newswire, telephone conversation, magazine and web text. It is tagged with eighteen entity types, some of which are: person, facility, organization, product and so forth. The dataset was assembled by \newcite{pradhan} for the CoNLL-2012 shared task, and specifies a standard train, validation, and test split followed in our evaluation.

\textit{KBP 2016}: The KBP 2016 trilingual EDL task \cite{kbp2016} consists of identifying named entities (including nested) from a collection of recent news article and discussion forum documents in three languages, and their classification to the following named and nominal entity types: person, geo-political entity, organization, location and facility. We use an in-house dataset that consists of 10k English and Chinese documents labelled manually using KBP 2016 format. Since KBP 2016 does not contain any training and development data, we use our in-house data as training and validation data with a 90:10 split. We also make use of the KBP 2015 dataset as additional data for training.

\textit{CoNLL-2002 SPA}: The CoNLL-2002  \cite{tjong2002} named entity data is tagged similarly to CoNLL-2003. We only make use of Spanish files for our experiments. 
\begin{table}
\centering
	\scalebox{0.8}{
	\begin{tabular}{  l l l l }\hline
		\textbf{Model} & \textbf{P} & \textbf{R} & \textbf{ F1} \\ \hline \hline
		\newcite{collobert2011} & $-$ & $-$ & $89.59$  \\
		\newcite{huang} & $-$ & $-$ & $90.10$ \\
		\newcite{strubell} & $-$ & $-$ &  $90.54$\\
		\newcite{yang} & $-$ & $-$ &  $\mathbf{90.94}$ \\ \hline
		\newcite{luo} \footnote{Numbers taken from the original paper} & $91.50$ & $91.40$& $91.2$  \\
		\newcite{lample}  & $-$ & $-$ & $90.94$ \\
		\newcite{chiu} & $91.39$ & $91.85$ & $\mathbf{91.62}$ \\ \hline
		\newcite{xu} & $93.29$ & $\mathbf{88.27}$ & $90.71$ \\
		\small{\newcite{xu} + dev set + 5-fold} & $92.58$ & $89.31$ & $\mathbf{90.92}$ \\ \hline
		Our model & $91.81$ & $\mathbf{89.85}$ & $\mathbf{90.82}$  \\
		Our model + dev set & $92.02$ & $90.30$ & $\mathbf{91.15}$\\\hline
	\end{tabular} }
	\caption{Results on the CoNLL-2003 ENG evaluation task. The three sections, in order, are models: trained with training set only, trained with both training and dev set, our baselines and our models.}
	\label{tbl:conll-results-eng}
\end{table}
\subparagraph*{Baselines}
Our baseline models are from \newcite{xu}. We use the author's findings for CoNLL-2003 and KBP 2016, and apply the implementation  released by the author to train the model with  OntoNotes 5.0 and CoNLL-2002 tasks.
\begin{table}
\centering
	\scalebox{0.8}{
	\begin{tabular}{ | l l l l |} \hline
		\textbf{Model} & \textbf{P} & \textbf{R} & \textbf{ F1} \\ \hline \hline
		\newcite{strubell} & $-$ & $-$ & $\mathbf{86.84}$ \\
		\newcite{chiu} & $86.04$ & $86.53$ &  $86.28$ \\
		\newcite{durrett} & $85.22$ & $82.89$ & $84.04$  \\ \hline
		\newcite{xu} & $86.84$ & $84.94$ & $85.88$ \\
		Our model & $86.95$ & $85.44$ & $\mathbf{86.19}$ \\\hline
	\end{tabular} }
	\caption{A comparison with the state-of-the-art results for the OntoNotes 5.0 ENG evaluation task. }
	\label{tbl:ontonotes-results-eng}
\end{table}
\begin{table}
\centering
	\scalebox{0.95}{
	\begin{tabular}{  | l l l l | } \hline
		\textbf{Model} & \textbf{P} & \textbf{R} & \textbf{ F1} \\ \hline \hline
		\newcite{che} & $74.38$ & $65.78$ & $69.82$ \\
		\newcite{pappu}  & $-$ & $-$ & $67.2$ \\ \hline
		\newcite{xu} & $72.91$ & $70.78$ & $71.83$ \\
		Our model & $76.20$ & $68.96$ & $\mathbf{72.40}$ \\\hline
	\end{tabular} }
	\caption{A comparison with published results for the OntoNotes 5.0 ZH evaluation task.}
	\label{tbl:ontonotes-results-cmn}
\end{table}
\begin{table}
\centering
	\scalebox{0.8}{
	\begin{tabular}{ | l l l l |} \hline
		\textbf{Model} & \textbf{P} & \textbf{R} & \textbf{ F1} \\ \hline \hline
		\newcite{santos} & $82.21$ & $82.21$ & $82.21$ \\
		\newcite{gillick} & $-$ & $-$ & $82.95$ \\
		\newcite{lample}  & $-$ & $-$  & $85.75$ \\
		\newcite{yang} & $-$ & $-$ &  $\mathbf{85.77}$ \\ \hline
		\newcite{xu} & $84.20$ & $82.26$ & $83.22$ \\
		Our model & $85.06$ & $82.31$ & $\mathbf{83.66}$ \\ \hline
	\end{tabular} }
	\caption{A comparison with the state-of-the-arts results for the CoNLL-2002 SPA evaluation task.}
	\label{tbl:conll-results-spa}
\end{table}
\section{Results and Discussion}
The results for the trilingual KBP 2016 task are presented in Table \ref{tbl:kbp2016}, where our system outperforms the baseline by 3.2 $F_1$ points for English and 4.3 $F_1$ points for Chinese. It also outperforms the best KBP 2016 English system by 1 $F_1$ point. It is worth considering that the best 2016 system uses 5-fold cross-validation.
The CoNLL-2003 results in Table \ref{tbl:conll-results-eng} show that our model is nearly on par with the state-of-the-arts compared to both models that used the dev-set to train the model, and to those who used training data only. The OntoNotes English and Chinese task results are presented in Tables \ref{tbl:ontonotes-results-eng} and \ref{tbl:ontonotes-results-cmn}, and the CoNLL-2002 results in Table \ref{tbl:conll-results-spa}. We do not observe significant improvement on the CoNLL-2002 task. Both of our baseline and proposed models outperform the other published results for the OntoNotes Chinese task.
\section{Related Work}
 In recent years, deep learning methods have gained much success in the NLP community. In view of the perceived limitations of FFNNs, recent methods involve more powerful neural networks, such as recurrent neural networks (RNNs), since they can process sequences of variable length \cite{huang}. As for character-level modeling, studies have turned to Convolutional neural networks (CNNs). \newcite{santos} have employed CNNs to extract character-level features for Spanish and Portuguese, and obtained successful NER results. \newcite{chiu} introduce a bi-directional LSTM-CNN architecture, and achieve state-of-the-art results in CoNLL2003. \newcite{strubell} uses Iterated dilated CNNs to reach state-of-the-art results in OntoNotes 5.0 English. Similarly, there have been conducted a few studies aiming at solving the problem of low recall in NER tasks \cite{mao, kuperus}.
\section{Conclusion}
We present a new neural model that can achieve near state-of-the-art results on a range of NER tasks by learning better representations of fragment and context features in NER systems. We use simple yet powerful neural networks with an effective context/fragment training approach. 

\newpage
\section{Supplemental Material}
\label{sec:supplemental}
\begin{table*}
	\centering
	\begin{tabular}{|l|c|c|c|c|c|}
		\hline
		\bf{Hyper-parameter}  & KBP ENG  & KBP CMN & KBP SPA  \\ \hline \hline
			Learning rate & 0.128 & 0.128 & 0.064 \\
			Fragment layers size & 512 & 512 & 412  \\
			Context layers size & 412 & 512 & 412  \\
			Shared layer size & 512 & 512 & 512  \\
      \hline
	\end{tabular}
	\caption{Hyper-parameters for the KBP 2016 experiments.}
	\label{tbl:kbp-hyper}	
\end{table*}

\begin{table*}
	\centering
	\begin{tabular}{|l|c|c|c|c|c|c|c|c|c|}
		\hline
		\bf{Hyper-parameter}  & CoNLL-2003 & OntoNotes ENG & OntoNotes ZH & CoNLL-2002 \\ \hline \hline
			Learning rate & 0.256 & 0.128 & 0.128 & 0.126 \\
			Fragment layers size  & 412 & 412 & 512 & 412 \\
			Context layers size  & 512 & 412 & 512 & 512 \\
			Shared  layer size & 512 & 612 & 512 & 512 \\
      \hline
	\end{tabular}
	\caption{Hyper-parameters for the CoNLL-2003, OntoNotes ENG/ZH and CoNLL-2002 experiments .}
	\label{tbl:hyperparameter}	
\end{table*}

\subsection{Hyper-parameters and Additional Info}
The hyper-parameters used in our experiments are shown in Tables \ref{tbl:kbp-hyper} and \ref{tbl:hyperparameter}. Most of the experiments use two layers dedicated to context/fragment features, and a single shared hidden layer. The exceptions are the English model for KBP 2016 which consists of three context layers, and the Chinese model for OntoNotes 5.0, which uses two shared layers. We use randomly initialized character embeddings of dimension 64. As specified in \newcite{xu}, Chinese is labelled at character level only.

\bibliography{emnlp2018}
\bibliographystyle{acl_natbib_nourl}

\end{document}